# Rethinking Urban Flood Risk Assessment By Adapting Health Domain Perspective

Zhewei Liu, Kai Yin*, Ali Mostafavi

UrbanResilience.AI Lab, Zachry Department of Civil and Environmental Engineering, Texas A&M University, DLEB 802B, College Station, TX, 77843

*kai_yin@tamu.edu

**Abstract**: Inspired by ideas from health risk assessment, this paper presents a new perspective for flood risk assessment. The proposed perspective focuses on three pillars for examining flood risk: (1) inherent susceptibility, (2) mitigation strategies, and (3) external stressors. These pillars collectively encompass the physical and environmental characteristics of urban areas, the effectiveness of human-intervention measures, and the influence of uncontrollable external factors, offering a fresh point of view for decoding flood risks. For each pillar, we delineate its individual contributions to flood risk and illustrate their interactive and overall impact. The three-pillars model embodies a shift in focus from the quest to precisely model and quantify flood risk to evaluating pathways to high flood risk. The shift in perspective is intended to alleviate the quest for quantifying and predicting flood risk at fine resolutions as a panacea for enhanced flood risk management. The decomposition of flood risk pathways into the three intertwined pillars (i.e., inherent factors, mitigation factors, and external factors) enables evaluation of changes in factors within each pillar enhance and exacerbate flood risk, creating a platform from which to inform plans, decisions, and actions. Building on this foundation, we argue that a flood risk pathway analysis approach, which examines the individual and collective impacts of inherent factors, mitigation strategies, and external stressors, is essential for a nuanced evaluation of flood risk. Accordingly, the proposed perspective could complement the existing frameworks and approaches for flood risk assessment.

1. **Introduction**

The core argument of this perspective paper is an alternate strategy of flood risk management in cities by drawing from and building upon analogies from health risk management approaches. The current approaches to flood risk management places greater emphasis on predicting external hazards and quantifying risks than on dissecting pathways that shape flood risk. Public health risk assessment recognizes the complexity of intertwined factors that shape health risk and focuses on understanding factors that shape predisposition to health risk (such as genetics), as well as interventions (such as lifestyle and diet) and external stressors (such as pollution and stress). This perspective focuses on characterizing pathways to risk rather than focusing on hazard drivers and a constant pursuit of risk estimation and quantification. Our perspective and ideas presented in this paper are not intended to undermine the value of the existing methods and frameworks for flood risk management. In fact, due to the complexity of flood risk analysis and different use cases of flood risk assessment outcomes, we argue that complementary frameworks and approaches are needed to characterize flood risk. Thus, our motivation here is to expand the breadth of perspectives and provide a fresh lens for understanding, analyzing, and managing flood risk.

The shift in perspective proposed in this paper is intended to alleviate the quest for quantifying and predicting flood risk at fine resolutions as a panacea for enhanced flood risk management. While quantifying and predicting flood risk is definitely useful for some use cases, such as insurance (pricing risk) and emergency response, the extent of computational effort and costs needed to obtain reliable flood risk maps usually make the process lengthy, costly, and more often than not, flood risk maps become outdated by the time they are available for use. The division of flood risk into the three pillars (i.e., inherent factors, mitigation factors, and external factors) can enable the evaluation of how changes in factors within each pillar would enhance and exacerbate flood risk to inform plans, decisions, and actions.

**Three-pillars Approach for Understanding Pathways to Flood Risk**

In the contemporary landscape of urban development, characterized by rapidly evolving environmental challenges, the management of flood risks stands out as a critical area of concern for city planners, policymakers, and citizens. Previous work on flood risk assessment have encompassed a range of analytical approaches to address various components of flood risk, as detailed in Table 1. These approaches include hydrological and hydraulic models that simulate the dynamics of floods, coupled with floodplain mapping techniques for identifying regions at risk. Additionally, probabilistic models and machine learning algorithms have been employed to forecast the likelihood and potential patterns of future flooding events. Geospatial analysis has been instrumental in pinpointing locations that are particularly vulnerable, while economic analysis has provided insights into the financial impacts of floods. Integrated approaches have been developed to synergize these diverse methods, offering a more comprehensive perspective on flood risk. Key tools, such as stage damage functions, damage matrices, and vulnerability indices have been utilized to quantify potential damages and assess the susceptibility of assets, facilitating a holistic assessment that encompasses flood hazard, exposure, vulnerability, and overall risk. All of these approaches to flood risk assessment are motivated by the quest for estimating and quantifying flood risk precision. While these approaches are highly valuable and have informed flood risk management practices and actions, they pay limited attention to pathways that create high flood risk. Using an analogy to health risk, while the quest for quantifying the level of risk of individuals to different diseases is worthwhile, a more practical approach is to understand pathways that lead to high risk of certain diseases and the adoption interventions that reduce health risk in individuals. Health risk literature has recognized the complexity of interactions among these three components and the difficulty of quantifying health risk likelihood of an individual based on the factors in each component. Due to this complexity, the focus in health risk literature has shifted to evaluating pathways to health risk rather than focusing on predicting and quantifying health risk at individual level. Such an approach is more informative for proactive actions to reduce health risk. However, the flood risk research has been on a quest to better predict external stressors and to quantify flood risk likelihood at fine spatial resolutions, which would be analogous to the aim for quantifying the likelihood of a certain disease in each individual.

**Table 1 Summarization of previous frameworks for flood risk rating**

| Aspect | Approaches | Description |
| --- | --- | --- |
| Flood hazard | Hydrological and hydraulic model | Simulate water movement and behavior, and use physical principles to predict how water interacts with the environment (Wang et al., 2019; Yin et al., 2023) |
| | Floodplain mapping | Use topographical data and historical flood information to identify flood-prone regions |
| | Probabilistic model | Calculate the likelihood of flooding based on statistical analysis of past flood events and environmental variables |
| | Machine learning algorithm predicting flood events | Use historical data to train algorithms that can predict future flood events, identifying patterns and trends that might not be evident through traditional methods (Lee et al., 2023; Liu et al., 2023a) |
| Flood exposure | Hydrological and hydraulic model | Similar to that of flood hazard analysis |
| | Geospatial analysis | Integrate spatial data, such as land elevation and land use, to identify areas vulnerable to flooding (Rajput et al., 2023; Yuan et al., 2023) |
| | Economic analysis | Assess the economic impacts of floods, help understand the financial implications of flood events and the value of prevention strategies (Farahmand et al., 2023; Wing et al., 2022) |
| | Integrated approaches | Combine multiple methodologies to provide a comprehensive understanding of flood exposure, and consider various aspects for holistic flood risk management (Yin and Mostafavi, 2023) |
| Flood Vulnerability | Stage damage functions | Quantify the relationship between water depth (stage) and the resultant damage to properties and infrastructure(Wing et al., 2020) |
| | Damage matrices | Associate different flood depths or durations with expected damage levels to various types of properties and assets (Sanders et al., 2022) |
| | Vulnerability indices | Numerical scores or ratings that summarize the susceptibility of an area or asset to flood damage (Fernandez et al., 2016) |
| Flood Risk | Hydrological & hydraulic Model | Similar to that in flood hazard analysis |
| | Combined estimation of hazard assessment, vulnerability analysis, and exposure | Integrate risk assessment by evaluating natural hazards, analyzing the vulnerability of communities or structures, and assessing the potential exposure to these risks (Yin and Mostafavi, 2023) |
| | Flood events prediction using probabilistic or machine learning model | Employ statistical or AI algorithms to predict flood occurrences based on historical data and environmental factors (Yuan et al., 2023) |

Recognizing this dichotomy in flood risk analysis, this perspective paper draws an insightful parallel between the methodologies employed in flood risk management and those in public health risk management. Public health risk assessment, a field renowned for its comprehensive and proactive stance, has long embraced a holistic view of managing health risks (Brownson, Fielding et al. 2009, Parkes, Poland et al. 2020). The primary health risk assessment approach encompasses a broad spectrum of factors, ranging from genetic predispositions, which determine an individual's inherent susceptibility to certain health conditions, to lifestyle choices, such as diet and exercise, which can significantly mitigate these risks (Harvey 2010, Sanderson, Waller et al. 2011). By paralleling these concepts, this paper seeks to illuminate the importance of considering the three intertwined pillars of flood risk to enable the analysis of pathways leading to high flood risk. This involves not only recognizing the inherent vulnerabilities of different urban areas, based on factors such as geography and infrastructure, but also understanding how mitigation interventions and external environmental changes interact to exacerbate or alleviate flood risks. Accordingly, this paper proposes a three-pillars approach in flood risk assessment, adopting a three-pronged model that mirrors key components of public health risk management: (1) inherent susceptibility, (2) mitigation strategies, and (3) external stressors, as illustrated in Figure 1.

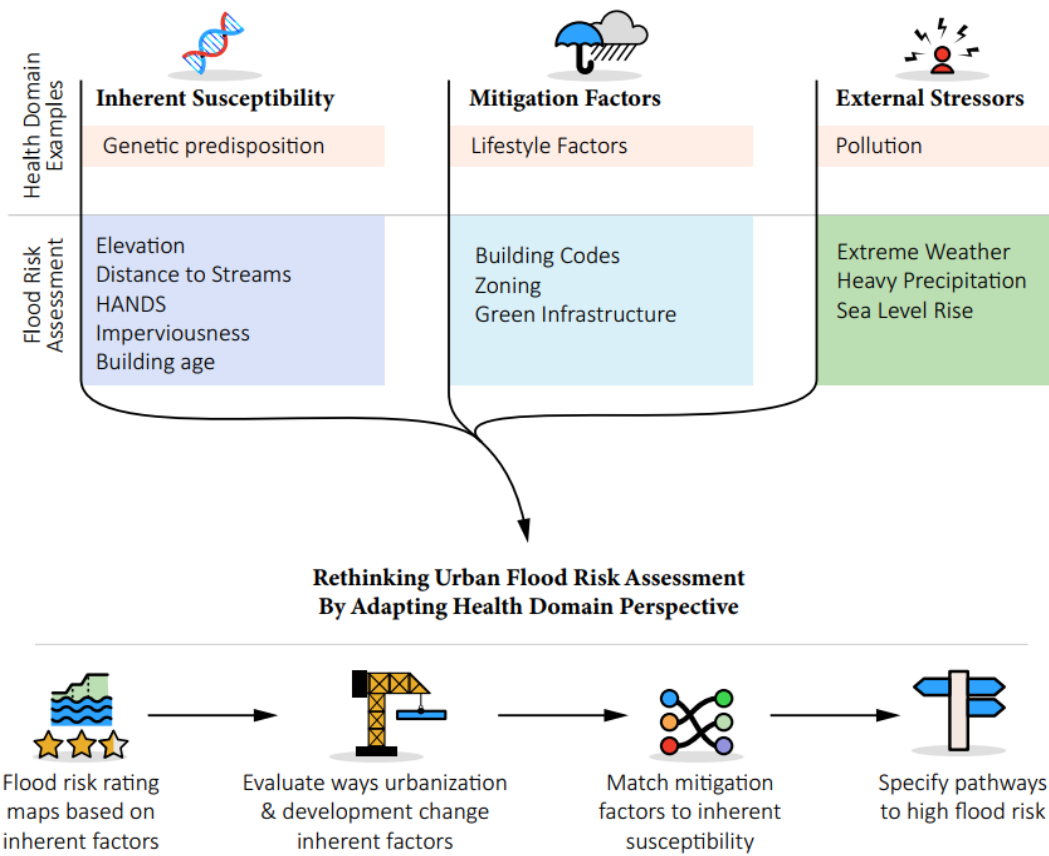

**Figure 1**. A novel approach for flood risk assessment, with consideration of three pillars that mirror health domain perspective: (1) inherent susceptibility, (2) mitigation factors, and (3) external stressors.

The first component, inherent susceptibility, is analogous to an individual's genetic makeup in health. Just as genetics can predispose individuals to certain health conditions, certain characteristics of an urban area, such as its geographical location, topography, soil composition, and existing infrastructure, can inherently make it more susceptible to flooding (Güneralp, Güneralp et al. 2015). Understanding these inherent characteristics is crucial for identifying areas at higher risk and necessitating targeted interventions. The second component involves mitigation strategies, which are equivalent to lifestyle choices in health management. In the realm of public health, lifestyle choices like diet, exercise, and preventive healthcare play a significant role in managing health risks (Chan and Woo 2010). Similarly, in flood risk management, proactive measures, such as low-impact development, implementation of building codes, development of green infrastructure like parks and rain gardens, and maintenance of stormwater management systems, function as crucial mitigating factors (Chan, Griffiths et al. 2018). These measures can significantly reduce the impact of floods in inherently susceptible areas. Finally, the three-pillars model considers external stressors, akin to environmental factors in public health. In health risk management, factors such as pollution, occupational hazards, and lifestyle stressors can exacerbate health risks. In a similar vein, flood risk management should contend with external stressors, such as climate change, sea-level rise, and increased frequency of extreme weather events (da Silva, Alencar et al. 2020). These stressors can intensify the frequency and severity of flooding, particularly in areas already vulnerable due to their inherent susceptibility and potentially inadequate mitigation strategies.

By evaluating these three pillars—inherent susceptibility, mitigation strategies, and external stressors—we can provide a more nuanced perspective to flood risk assessment. This perspective not only acknowledges the complexity of the factors at play but also offers a framework for targeted and proactive interventions, drawing valuable parallels from the field of public health. The following sections 2 to 4 elaborate each of the three pillars for our flood risk assessment models. For each component, we discuss the analogy between the concepts borrowed from health risk assessment and provide illustrative examples to enhance understanding. In section 5, we introduce a holistic approach for identifying regional flood risks, achieved through a comprehensive consideration of these three key components. Finally, we summarize the model's contributions and explore the opportunities associated with effectively applying our proposed model in flood risk assessment scenarios.

## 2. Inherent Flood Risk Susceptibility

The first pillar of the model is the concept of inherent susceptibility, offering an analogy to the concept of genetic predisposition in individual health. This comparison is particularly apt because, in both cases, the inherent characteristics significantly determine the likelihood and potential severity of future risks, either in health-related or environmental fields. Just as certain genetic profiles can predispose individuals to specific health conditions, various characteristics of an urban area can similarly render it inherently more susceptible to flooding. These characteristics, much like genes in the human body, are intrinsic to the very nature of the urban landscape

The geographical location of an urban area is one of these critical components. Cities located near coastlines, rivers, or in low-lying areas are naturally more prone to flooding. Coastal cities, for instance, face risks from storm surges and sea-level rise, while cities near rivers can experience flooding from river overflow. Low-lying areas, even those not immediately adjacent to water sources, are susceptible to flash floods during heavy rainfall due to their positioning as natural catchment zones (Helderop and Grubesic 2019). Topography, or the physical layout of the land, also significantly influences how water flows and accumulates in an urban area. Hilly or uneven terrains can result in runoff accumulating in specific areas, while flat regions, particularly densely built-up ones, often lack natural routes for water drainage. Soil composition further adds to this complex picture (McGrane 2016). Urban areas with soil with a high clay content have poor water absorption capacity, leading to quicker runoff and increased flooding potential. Conversely, areas with sandy or loamy soils allow better water infiltration, potentially mitigating surface flooding but perhaps dealing with groundwater level challenges during prolonged wet periods (Yang and Zhang 2015). The design and condition of existing urban infrastructure are equally important factors. Aging or inadequate drainage systems and poorly maintained levees or flood barriers significantly increase flood risks (Sohn, Brody et al. 2020). The presence of impervious surfaces like concrete and asphalt in urban settings further exacerbates these risks by preventing natural water absorption and contributing to runoff.

Understanding and managing these inherent urban characteristics is fundamental in effectively addressing urban flood risks. Just as in healthcare, where understanding a patient's genetic makeup can guide more personalized and effective treatment plans, in urban planning, a deep understanding of a city's inherent susceptibility to floods can lead to more tailored and effective flood management strategies. This can involve specific urban design and infrastructure choices that mitigate the risks associated with these inherent characteristics. This knowledge is critical in developing targeted interventions in high-risk areas, such as reinforcing flood defenses, modifying land use planning, and improving infrastructure to mitigate flood risks. Once areas of high risk are identified, targeted flood risk management strategies can be developed and implemented. By comprehensively evaluating factors such as geographical location, topography, soil composition, and existing infrastructure, urban planners can devise targeted and effective strategies to mitigate flood risks, ensuring the safety and resilience of urban populations against the challenges of flooding.

3. **Mitigation Factors in Flood Risk Management**

The second pillar for flood risks assessment is mitigation factors, which hold a parallel position to lifestyle choices in public health. This analogy is particularly insightful when we consider how individual decisions and actions can profoundly influence outcomes.

In health risk management, the choices one makes in terms of diet, exercise, and preventive healthcare can have a substantial impact on their capacity to deal with health challenges. Similarly, in the realm of flood risk management, proactive and preventive measures serve as the cornerstone for effectively reducing the impact of floods. These measures can be seen as the "lifestyle choices" for communities for flood risk management. Just as a nutritious diet strengthens the body against illness, thoughtful urban planning that accommodates natural waterways, preserves floodplains, and avoids overdevelopment in high-risk areas can significantly reduce the vulnerability of cities to flood events (Firehock and Walker 2015; Zhang, et al., 2022). Particularly, there are some key useful mitigation strategies for impacting flood risks:

- **Effective urban planning**: Urban planning plays a crucial role in flood risk mitigation. This involves thoughtful land-use decisions, such as avoiding construction in high-risk areas and preserving natural floodplains and wetlands that act as natural barriers against floods.
- **Implementation of building codes**: Strengthening building codes to require flood-resistant materials and designs is vital in flood-prone areas. This can include elevating structures, using water-resistant building materials, and designing buildings to withstand flood pressures.
- **Development of green infrastructure**: Implementing green infrastructure, such as parks, rain gardens, and green roofs, serves multiple purposes (Wang, et al., 2020). These spaces serve to absorb and hold stormwater and significantly reduce flood risk.
- **Maintenance of stormwater management systems**: Regular maintenance and upgrading of stormwater management systems ensure they function effectively, particularly during heavy rainfall events. This includes cleaning drainage systems, repairing levees, and installing advanced technologies for water management.

Similar to lifestyle changes in health risk management, mitigation strategies in flood risk management are inherently dynamic and require continuous evolution to remain effective. This dynamism is primarily driven by some key factors, such as technological advancements and changes in urban landscapes:

- **Technological advancements continually redefining what is possible in flood risk management**: The development of new materials, construction techniques, and data analysis tools allows for more effective and efficient flood defenses. Innovations in areas such as emerging technologies in communication and information dissemination, geographic information systems (GIS), and machine learning provide urban planners with more precise data and predictive models, enhancing their ability to identify flood-prone areas, to raise public awareness, and to design effective mitigation strategies.
- **Changes in urban landscapes also necessitate the adaptation of mitigation strategies:** As cities expand and evolve, their hydrological characteristics change.

> New developments can alter the natural flow of water, create additional impervious surfaces, and put pressure on existing drainage and water management systems. These changes can increase the risk of flooding or shift the most at-risk areas. Mitigation strategies, therefore, must be reassessed and modified to account for these developments. This could involve redesigning or upgrading infrastructure, implementing new urban planning guidelines, or restoring natural floodplains that have been affected by urbanization (Yao, et al., 2021).

By regularly reassessing and updating these strategies in response to technological advancements and changes in urban environments, urban planners and policymakers can better safeguard communities against the dynamic and increasing challenges of urban flooding, which is key to building and maintaining resilient, flood-resistant urban spaces.

### 4. External Stressors: Amplifiers of Flood Risk

The final component of our model for flood risk management is the consideration of external stressors, akin to environmental factors in public health. Just as various external factors, such as air pollution, occupational hazards, and lifestyle stressors, can amplify health risks in individuals, similar elements play a significant role in escalating flood risks in urban environments (Wang, et al., 2022; Liu, et al., 2023b).

Translating this analogy to flood risk management, urban areas face a series of external stressors that can dramatically heighten the risk and impact of flooding. Primary among these stressors is climate change, an overarching phenomenon causing widespread shifts in weather patterns. Climate change is responsible for more frequent and severe weather events, such as heavier than usual rainfall, hurricanes, and typhoons, all of which directly contribute to increased flooding risks (Van Aalst 2006). The unpredictability and increased frequency of extreme weather events (e.g., atmospheric rivers, hurricanes, and torrential storms) poses a significant challenge to effective flood risk management (Field 2012). Urban areas around the world are increasingly experiencing bouts of intense rainfall, causing flash floods and urban flooding. These events can cause substantial damage to infrastructure, disrupt essential services, and lead to significant economic and social disruptions. Urban areas that already have inherent susceptibility factors, such as poor drainage systems or a high percentage of impervious surfaces, are particularly vulnerable to these extreme events.

Another critical external stressor in the context of flood risk is sea-level rise, a direct consequence of global warming (Brown, Nicholls et al. 2013). Rising sea levels pose a persistent threat to coastal cities, exacerbating the risk of coastal flooding and storm surges. This challenge is especially daunting for low-lying urban areas, where even small rises in sea levels can lead to significant flooding issues. Coastal erosion, saltwater intrusion, and the loss of habitat caused by rising sea levels further compound the problem, necessitating urgent and innovative responses in urban planning and flood management.

Understanding and adapting to these external stressors requires a multi-faceted approach that involves not only addressing the inherent and mitigation factors within urban areas but also actively adapting to and mitigating the broader impacts of these external challenges. This could involve strategies ranging from reducing the inherent susceptibility of communities or implementing mitigation measures.

5. **Pathways to Flood Risk**

The main benefit of the proposed three-pillars model is to enable a shift in focus from the quest to quantify the probability of external stressors (e.g., return period of heavy precipitation events) to analyzing pathways to reduce flood risk and place a greater focus on holistically reducing the inherent susceptibilities and implementing mitigation measures. As with an individual hoping to prevent heart disease or cancer, the best approach is to understand genetic predispositions and make lifestyle changes to reduce the risk of diseases. By evaluating inherent susceptibility factors and mitigation strategies, a clear picture emerges of the baseline risk and the effectiveness of existing measures to counteract this risk. This step is crucial for understanding the fundamental vulnerability of an area to flooding. Inherent factors, such as geographical location and urban infrastructure, set the stage for the degree of flood susceptibility exists in an area. Mitigation factors, including urban planning, land-use policies, and infrastructure developments, act to modify this inherent risk.

The effects of external stressors can be viewed in conjunction with the inherent susceptibility and mitigation measures forming pathways to flood risk. For instance, an area characterized by high inherent susceptibility and inadequate mitigation efforts can rapidly escalate to high-risk status under moderate external stressors like a 20-inch cumulative rainfall. Conversely, an area with low inherent susceptibility, bolstered by moderate mitigation efforts, may generally be considered at lower risk. However, this area's risk level could escalate to moderate or high under severe external stressors, such as a 40-inch rainfall event. Such pathway perspectives can lead to the development of new tools for communicating flood risk. For example, instead of 100- or 500-year flood plains, maps can be created to determine the flood risk of different areas of communities based on the factors related to inherent susceptibility. These baseline risk maps can be supplemented with additional information regarding what level of external stressor (e.g., cumulative and peak rainfall) would yield a high-risk pathway in an area. Such tools could complement the existing methods for capturing and communicating flood risk. Also, the baseline flood risk rating maps based on inherent susceptibility could be updated based on urban growth, development, or implementation of mitigation measures to continuously evaluate and monitor the evolution of inherent susceptibility to flood risk in different areas of communities. Such evaluation and monitoring would create a positive feedback loop for urban planning, flood mitigation, and adaptation. Hence, this approach facilitates the identification of specific pathways to high flood risk, enabling targeted interventions. By understanding how different extents of external stressors interact with inherent susceptibility and mitigation efforts, urban planners and policymakers can more effectively allocate resources, design infrastructure, and plan urban development to mitigate flood risks.

6. **Concluding Remarks**

This paper presents a new perspective for flood risk assessment with an analogy borrowed from health risk management consisting of three intertwined pillars: (1) inherent susceptibility (2) mitigation strategies, and external stressors. These intertwined pillars represent pathways to flood risk levels in communities. The presented perspective and the proposed model offer refreshing contributions and implications for research and practice of flood risk assessment to inform future research.

Based on the three-pillars framework, the following ideas can be evaluated to enhance the understanding and management of flood risk in cities:

- **Flood risk rating maps based on inherent factors**: Areas of cities can be classified into flood risk categories based on their inherent factors irrespective of external stressors. Simple index-based methods or advanced machine learning models could be adopted to rate the inherent susceptibility of an area to flood risks. This approach would enable frequent and relatively easier and faster updating of maps based on changes in the inherent factors. For example, if development reduces imperviousness in an area, the rating of the area would change.

- **Evaluate ways urbanization and development change inherent factors**: Like the human body, urban systems are continuously evolving, thus changing the inherent flooding risk factors. Using the three-pillars framework, we can better examine and communicate ways urbanization and development patterns would change the inherent susceptibility of urban areas to flood risk and devise integrated urban design strategies in which the assessment and mitigation of flood risk is integrated into urban development plans and policies.
- **Match mitigation factors to inherent susceptibility**: Based on flood susceptibility rating on different areas and the underlying inherent factors, suitable mitigation factors could be selected. Similarly, scores can be developed to quantify the extent of flood mitigation factors in a spatial area given its inherent susceptibility
- **Specify pathways to high flood risk**: With the inherent factors and mitigation factors evaluated separately, we can now examine the overall flood risk of an area for different extents of external stressors. For example, an area with high inherent susceptibility and low mitigation factors with a moderate to low external stressor (e.g., cumulative rainfall of 20 inches) would have a high flood risk. Or an area with low inherent susceptibility and moderate mitigation factors could still have moderate to high flood risk if external stressors exceed a certain level (e.g., 40 inches of rainfall). While such evaluation of pathways to high flood risk is ostensibly less accurate and quantitative than advanced hydrological and hydraulic models, this approach can derive actions in the absence of advanced and updated models.

By adopting a pillars-based and pathway-oriented model akin to those used in public health—which consider inherent susceptibilities, mitigation strategies, and external stressors—urban planners and policymakers can develop more resilient and adaptable flood management systems. This approach involves ongoing monitoring and assessment of urban areas to preemptively strengthen flood-prone zones, paralleling public health campaigns, and enabling residents to actively participate in flood mitigation efforts. This strategy not only improves the characterization of flood risk but also bolsters integrated urban design strategies in flood risk management.